\documentclass[10pt,twocolumn,letterpaper]{article}

\usepackage[pagenumbers]{cvpr} 

\usepackage{graphicx}
\usepackage{amsmath}
\usepackage{amssymb}
\usepackage{booktabs}
\usepackage{color, colortbl}
\usepackage[accsupp]{axessibility}
\usepackage{multirow}
\usepackage{makecell}
\newcommand{\keypoint}[1]{\vspace{0.01cm}\noindent\textbf{#1}\quad}
\newcommand{\cut}[1]{}
\definecolor{Gray}{gray}{0.9}
\usepackage[pagebackref,breaklinks,colorlinks]{hyperref}

\usepackage[capitalize]{cleveref}
\crefname{section}{Sec.}{Secs.}
\Crefname{section}{Section}{Sections}
\Crefname{table}{Table}{Tables}
\crefname{table}{Tab.}{Tabs.}

\def\cvprPaperID{5825} 
\def\confName{CVPR}
\def\confYear{2023}

\begin{document}

\title{What Can Human Sketches Do for Object Detection?\\[-0.3cm]}

\author{Pinaki Nath Chowdhury \hspace{.2cm}  Ayan Kumar Bhunia \hspace{.2cm} Aneeshan Sain \hspace{.2cm}  Subhadeep Koley \\
Tao Xiang\hspace{.3cm}  Yi-Zhe Song \\
SketchX, CVSSP, University of Surrey, United Kingdom.  \\
{\tt\small \{p.chowdhury, a.bhunia, a.sain, s.koley, t.xiang, y.song\}@surrey.ac.uk} \vspace{-0.3cm}
}

\maketitle

\begin{abstract}
Sketches are highly expressive, inherently capturing subjective and fine-grained visual cues. The exploration of such innate properties of human sketches has, however, been limited to that of image retrieval. In this paper, for the first time, we cultivate the expressiveness of sketches but for the fundamental vision task of object detection. The end result is a sketch-enabled object detection framework that detects based on what \textit{you} sketch -- \textit{that} ``zebra'' (e.g., one that is eating the grass) in a herd of zebras (instance-aware detection), and only the \textit{part} (e.g., ``head" of a ``zebra") that you desire (part-aware detection). We further dictate that our model works without (i) knowing which category to expect at testing (zero-shot) and (ii) not requiring additional bounding boxes (as per fully supervised) and class labels (as per weakly supervised). Instead of devising a model from the ground up, we show an intuitive synergy between foundation models (e.g., CLIP) and existing sketch models build for sketch-based image retrieval (SBIR), which can already elegantly solve the task -- CLIP to provide model generalisation, and SBIR to bridge the (sketch$\rightarrow$photo) gap. In particular, we first perform independent prompting on both sketch and photo branches of an SBIR model to build highly generalisable sketch and photo encoders on the back of the generalisation ability of CLIP. We then devise a training paradigm to adapt the learned encoders for object detection, such that the region embeddings of detected boxes are aligned with the sketch and photo embeddings from SBIR. Evaluating our framework on standard object detection datasets like PASCAL-VOC and MS-COCO outperforms both supervised (SOD) and weakly-supervised object detectors (WSOD) on zero-shot setups. Project Page: \url{https://pinakinathc.github.io/sketch-detect}

\end{abstract}

\vspace{-0.5cm}
\section{Introduction}
\label{sec:intro}

\begin{figure}
    \centering
    \includegraphics[width=\linewidth]{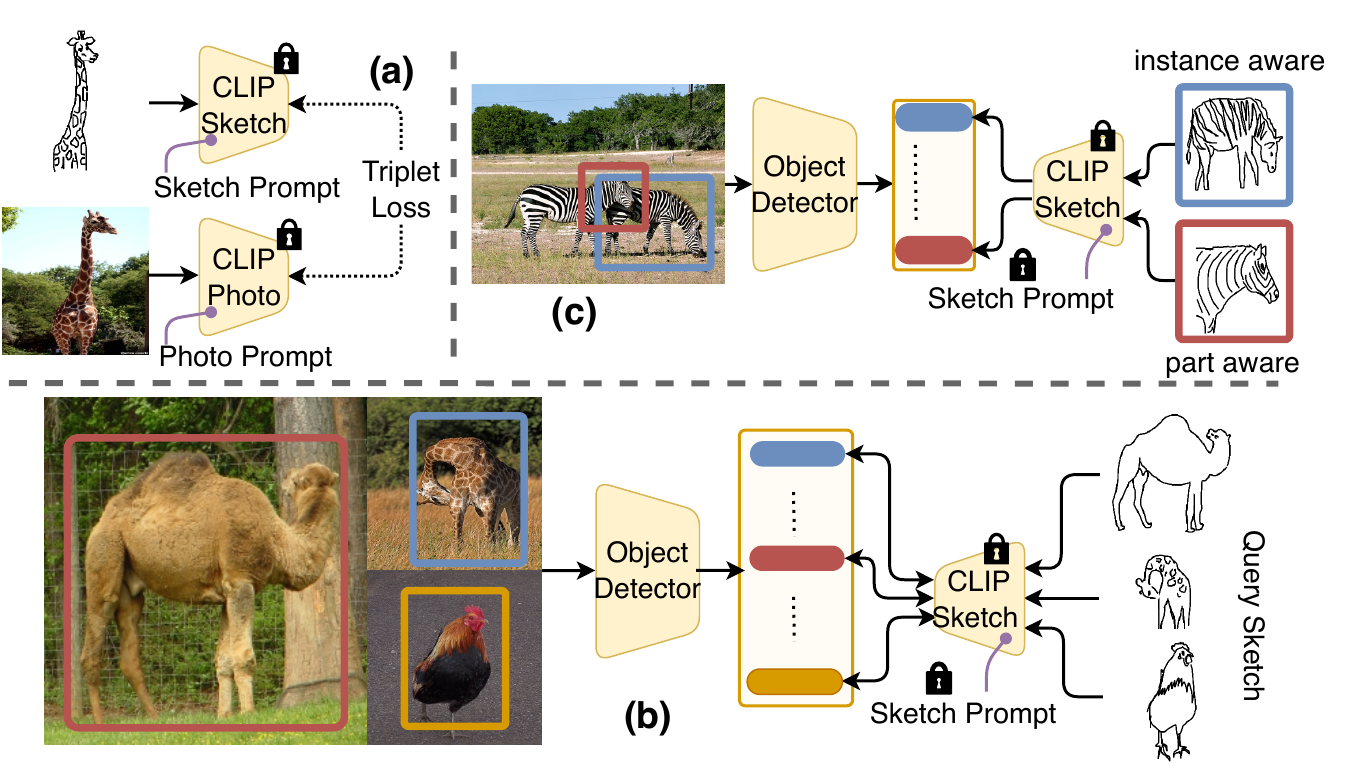}
    \vspace{-0.9cm}
    
    \caption{We train an object detector using SBIR models. (a) First, we train an FG-SBIR model using existing sketch--photo pairs that generalise to unseen categories. (b) To train the object detector module, we tile multiple object-level photos from SBIR datasets \cite{sketchy} and use its paired sketch encoding via a pre-trained sketch encoder to align the region embedding of detected boxes. (c) Inclusion of sketches for object detection opens several avenues like detecting a specific object for query sketch (e.g., detect a ``zebra" eating grass) or part of an object (e.g., ``head" of ``zebra").}
    \label{fig:problem-setup}
    \vspace{-0.4cm}
\end{figure}

Sketches have been used from prehistoric times for humans to express and record ideas \cite{hertzmann2020perception, sayim2011sketch}. The level of expressiveness \cite{kennedy1974sketch, goodwin2007} they carry remains unparalleled today even in the face of language \cite{text-sketch-dilemma, fscoco} -- recall that moment that you want to resort to pen and paper (or Zoom Whiteboard) to sketch down an idea?

Sketch research has also flourished over the past decade \cite{liveSketch, wang2022medical, sketchformer, yelamarthi2018sketch}, with a whole spectrum of works on traditional tasks such as classification \cite{sketchrnn2018} and synthesis \cite{chowdhury20223Dsynthesis, cusuh2022synthesis, DeepFaceVideoEditing2022}, and those more sketch-specific such as modelling visual abstraction \cite{umar2018abstraction, alaniz2022primitives}, style transfer \cite{styleMeUp} and continuous stroke fitting \cite{sketchODE}, to cute applications such as turning a sketch into a photo classifier \cite{sketch-a-classifier, bhunia2022incremental}. 

The expressiveness of sketches, however, has been only explored in the form of sketch-based image retrieval (SBIR) \cite{doodle-to-search, sketchformer, sketchmate}, especially the fine-grained \cite{bhunia2022worrying, bhunia2020sketch, bhunia2021semi} variant (FG-SBIR). Great strides have been made, with recent systems already reaching maturity for commercial adaptation \cite{bhunia2022worrying} -- a great testimony to how cultivating sketch expressiveness can make a real impact. 

In this paper, we ask the question -- what can human sketches do for the fundamental vision tasks of object detection? The envisaged outcome is, therefore, a sketch-enabled object detection framework that detects based on what you sketch, i.e., how \textit{you} want to express yourself. Sketching a ``zebra eating the grass" (in \cref{fig:problem-setup}) should detect ``that" zebra from a herd of zebras (instance-aware detection), \textit{and} it will also give you the freedom to be specific with parts (part-aware detection), so if the ``head" of a ``zebra" is what you would rather desire, then just sketch the very head. 

Instead of devising a sketch-enabled object detection model from the ground up, we show that an intuitive synergy between foundation models (e.g., CLIP \cite{CLIP}) and off-the-shelf SBIR models \cite{yu2016shoe, bhunia2022adaptive} can already, rather elegantly, solve the problem -- CLIP to provide model generalization, and SBIR to bridge the (sketch$\rightarrow$photo) gap. In particular, we adapt CLIP to build sketch and photo encoders (branches in a common SBIR model) by learning independent prompt vectors \cite{maple2022, sain2023clip} separately for both modalities. More specifically, during training, the learnable prompt vectors are prepended into the input sequence of the first transformer layer of CLIP's ViT backbone \cite{ViT} while keeping the rest frozen. As such, we inject model generalization into the learned sketch and photo distributions. Next, we devise a training paradigm to adapt the learned encoders for object detection, such that the region embeddings of detected boxes are aligned with the sketch and photo embeddings from SBIR. This allows our object detector to train \textit{without} requiring additional training photos (bounding boxes and class labels) from auxiliary  datasets.

To make our sketch-based detector more interesting (general-purpose \cite{minderer2022simple, chowdhury2023scenetrilogy}), we further dictate it also works in a zero-shot manner. For that, following \cite{wsddn}, we extend object detection from a pre-defined fixed-set setup to an open-vocab setup. Specifically, we replace the classification heads in object detectors with prototype learning \cite{prototype-segmentation-2021}, where the encoded query sketch features act as the support set (or prototypes). Next, the model is trained under the weakly supervised object detection (WSOD) setting \cite{wsddn, wsodsurvey2022}, using a multi-category cross-entropy loss over the prototypes of all possible categories or instances. However, while SBIR is trained using object-level (single object) sketch/photo pairs, object detection works on image-level (multiple categories). Hence, to train object detectors using SBIR, we also need to bridge the gap between object and image-level features. Towards this, we use a data augmentation trick that is embarrassingly simple yet highly effective for robustness towards corruption and generalisation to out-of-vocab \cite{cutmix, mixup} -- we randomly select $n=\{1, \dots, 7\}$ photos from SBIR datasets \cite{sketchrnn2018, sketchy} and arbitrarily tile them on a blank canvas (similar to CutMix \cite{cutmix}).

In summary, our contributions are (i) for the first time cultivating the expressiveness of human sketches for object detection, (ii) sketch-based object detector that detects what you intend to express in your sketch, (iii) an object detector that is both instance-aware and part-aware, in addition to performing conventional category-level detection. (iv) a novel prompt learning setup to marry CLIP and SBIR to build the sketch-aware detector that works without needing bounding box annotations (as supervised \cite{faster-rcnn}), class labels (as weakly supervised \cite{wsddn}), and in a zero-shot manner. (v) results outperform both supervised (SOD) and weakly supervised object detectors (WSOD) on zero-shot setup.


\section{Related Works}
\label{sec:related-works}

\keypoint{Sketch for Visual Understanding}
Hand-drawn sketches serve as a useful query modality for visual understanding tasks that involve human perception and structural cues. Sketches not only convey a visual description \cite{hertzmann2020perception} but also exhibit artistic styles \cite{zhang2021smartshadow}. This makes sketch a vital querying modality for the creative industry, like artistic image editing \cite{yang2020surgery} and animation \cite{xing2015autocomplete}. Unlike photos that are passively captured by a camera, sketches are actively drawn by humans, which makes them a good visual representation \cite{pixelor, bhunia2023sketch2saliency} enriched with human participation. Apart from the widely explored sketch-based image retrieval \cite{bhunia2021semi, bhunia2020sketch, styleMeUp, sketchformer}, sketch as a query has shown potential in several vision understanding tasks like incremental learning \cite{bhunia2022incremental} image and video synthesis \cite{cusuh2022synthesis, DeepFaceVideoEditing2022, koley2023picture}, representation learning \cite{alaniz2022primitives, clipasso}, image-inpainting \cite{xie2021inpainting}, 3D shape retrieval \cite{xu20223Dretrieval}, 3D shape modelling \cite{chowdhury20223Dsynthesis}, medical image analysis \cite{wang2022medical, kobayashi2023medical}, object localisation \cite{tripathi2020object, riba2021object} and segmentation \cite{hu2020segmentation, qi2022segmentation}.

Studying sketch as a query for object detection by Tripathi \etal \cite{tripathi2020object}, several limitations surfaced with respect to problem definition as well as architectural designs. 
Firstly, instead of fine-grained matching, sketch was used to specify object category (easier via text/keyword \cite{CLIP, gu2022open-vocab-OD}), thus overlooking the potential of sketch to model fine-grained details. Secondly, it requires both bounding-box and sketch annotation, which increases the annotation budget without significant improvement in performance over traditional object detection setups. Thirdly, due to an expensive annotation, only fewer than $50\%$ object categories in existing object detection datasets \cite{pascalVOC, mscoco} are available for training. Finally, using an early fusion strategy \cite{xu2022MML} of sketch with object detection results in recomputing object regions for each new sketch -- leading to a slower detection framework with increasing query sketches. In this paper, we propose a fine-grained sketch-based object detection framework that uses only object-level sketch photo pairs without any bounding-box annotations for training and is scalable with multiple fine-grained query sketches, even under zero-shot setup.

\keypoint{Supervised Object Detection} 
Object detection jointly localises and identifies objects in an image. Traditional object detectors rely on large datasets such as PASCAL VOC \cite{pascalVOC} and MS-COCO \cite{mscoco}, containing thousands of examples per object category which are quite time-consuming to annotate, unlike our pipeline, that leverage sketch-photo pairs as annotation. 
Existing literature on object detection is bifurcated as: (i) fast yet less accurate single-shot \cite{yolo, yolov3, SSD, lin2017focal-loss, centerNet}; (ii) slow but more accurate two-stage object detectors \cite{rcnn, fast-rcnn, faster-rcnn, maskrcnn}. To fully exploit the fine-grained cues provided by sketch, our proposed method aligns with the accurate two-stage detectors that predict object regions using selective search in RCNN \cite{rcnn}, ROI pooling \cite{fast-rcnn} in Fast-RCNN, and Region Proposal Network (RPN) with ROIAlign in Faster-RCNN \cite{faster-rcnn}. While there has been several attempts with sophisticated architectural modifications   \cite{law2018cornerNet, zhou2019grouping, centerNet}, Faster-RCNN \cite{faster-rcnn} still acts a fundamental building block for several downstream tasks like scene-graph generation \cite{yang2018scenegraph}, visual grounding \cite{mouzenidis2021grounding}, and relationship prediction \cite{zhu2018relationship}. Therefore, we resort to the more traditional Faster-RCNN based two-stage pipeline.

\keypoint{Weakly Supervised Object Detection (WSOD)} Collecting bounding box annotation per object category is already a time-consuming process, which is aggravated even further by fine-grained object detection (e.g., recognising animal species). To overcome this, existing WSOD adopt two schools of thought: (a) formulate this as a multiple instance learning (MIL) \cite{dietterich1997MIL, wsddn, li2016wsod, diba2017wsod, jie2017wsod, zhang2018wsod, tang2018wsod, shen2019wsod} problem that interpret an image as a bag of proposals or regions. The image is labelled positive if one of the regions tightly contains the object of interest; otherwise negative. (b) CAM-based methods \cite{zhou2016CAM, zhang2018cam} that use class activation maps to predict proposals. Specifically, an image is fed to a backbone network to generate a feature vector from which the bounding box of each class is predicted by thresholding activation maps with the highest probability. Although CAM-based methods are faster, we use MIL-based technique as it can detect \textit{multiple instances} within the \textit{same} category \cite{wsodsurvey2022}.

\keypoint{Data Augmentation in Computer Vision} Data augmentation improves the sufficiency and diversity of training data. Approaches vary from simple image rotation and flipping to more advanced techniques of image erasing \cite{gridmask} like CutOut \cite{devries2017cutout}, Hide-and-Seek \cite{hide-and-seek} and image mixing like MixUp \cite{mixup} and {CutMix} \cite{cutmix}. 
Aiming to generalise sketch-based object detection to complex scenes while training exclusively on existing object-level sketch photo pairs \cite{sketchy}, we employ a CutMix \cite{cutmix} like data augmentation trick -- a method that replaces removed sub-regions with a patch from another image to synthesise new images.

\keypoint{Sketch-Based Object Representation}
Sketch with its intrinsic ability to model fine-grained visual details makes it an ideal modality  for object retrieval, giving rise to avenues like \emph{category-level} \cite{liveSketch, yelamarthi2018sketch, doodle-to-search, sketchformer, sketchmate} and fine-grained (FG) \emph{instance-level} \cite{sain2023exploiting, bhunia2022worrying, bhunia2020sketch, bhunia2021semi} sketch-based image retrieval (SBIR). Contemporary research also explored zero-shot SBIR \cite{doodle-to-search, yelamarthi2018sketch, sain2022sketch3t}, cross-domain translation \cite{kaiyue2017cross} and approaches like reinforcement learning based on-the-fly retrieval \cite{bhunia2020sketch}, self-supervised \cite{pang2020jigsaw, vector-raster}, etc. Apart from object-level images, retrieving sketched objects from scene images was studied using graph convolutional networks \cite{liu2020scenesketcher} and optimal transport \cite{partially-does-it}. Similar to cross-category FG-SBIR \cite{kaiyue2017cross, bhunia2022adaptive}, here we explore fine-grained sketch photo association for object detection by adapting large vision-language models like CLIP \cite{CLIP} using prompt engineering \cite{zhou2022visualprompt}.

\section{Proposed Method}
\label{sec:proposed-method}
\vspace{-0.2cm}

\keypoint{Overview} We propose a new paradigm training object detection without bounding box annotation or image-level class labels. Instead, we use sketch-based image retrieval for supervision. This leads to several emergent behaviours (i) fine-grained object detection -- specify focused region-of-interest using fine-grained visual cues in sketch. (ii) category-level object detection -- specify the category of detected instances via sketch. (iii) part-level object detection -- detect specified parts (e.g., ``head" and ``legs" of a ``horse").

\subsection{Background}\label{sec: background}
\vspace{-0.1cm}

Our framework has two key modules -- Object Detection and Sketch-Based Image Retrieval (category-level and fine-grained). For completeness, we give a brief background.

\vspace{-0.3cm}
\begin{figure}[!h]
    \centering
    \includegraphics[width=0.8\linewidth]{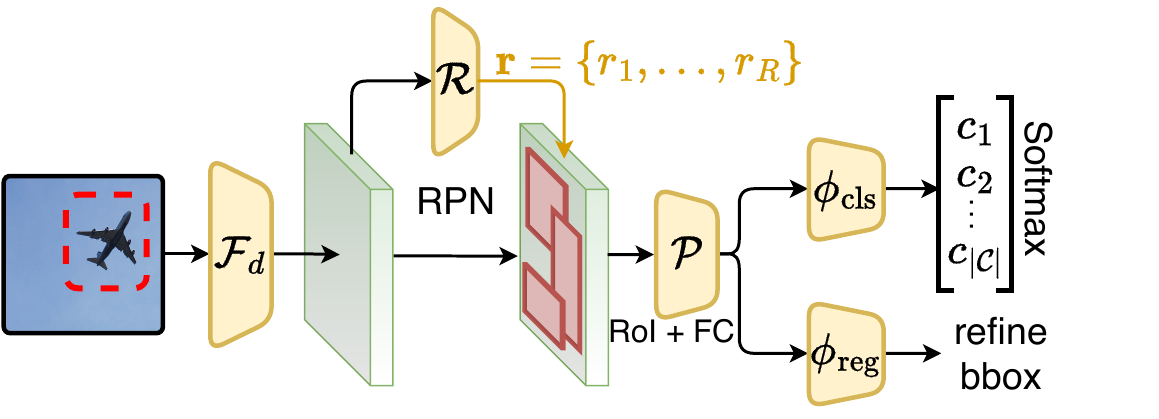}
    \vspace{-0.4cm}
    \caption{Faster-RCNN \cite{faster-rcnn} use image encoder $\mathcal{F}_{d}$ and RPN $\mathcal{R}$ to generates box proposals. Feature maps of proposals, computed via RoI pool $\mathcal{P}$, predicts class probabilities and box regression.}
    \label{fig: faster-rcnn}
    \vspace{-0.1cm}
\end{figure}

\keypoint{Baseline Supervised Object Detection} We briefly introduce a supervised object detection (SOD) framework, Faster-RCNN \cite{faster-rcnn} that remains state-of-the-art \cite{yang2018scenegraph, mouzenidis2021grounding, zhu2018relationship}. Given a photo $\mathbf{p} \in \mathbb{R}^{H \times W \times 3}$, a backbone feature extractor (VGG \cite{vgg}, or ResNet \cite{he2015resnet}) $\mathcal{F}_{d}(\cdot): \mathbb{R}^{H \times W \times 3} \rightarrow \mathbb{R}^{H' \times W' \times 512}$ computes feature map $f_{\mathbf{p}} \in \mathbb{R}^{H' \times W' \times 512}$. Next, a two-stage process is followed: (i) Given backbone feature map $f_{\mathbf{p}}$, a region proposal network (RPN) $\mathcal{R}: \mathbb{R}^{H' \times W' \times 512} \rightarrow \mathbb{R}^{R \times 5}$ generates rectangular boxes (i.e., proposals) $\mathbf{r} = \{r_1, \dots, r_R\}$, where $\mathbf{r} \in \mathbb{R}^{R \times 4}$ and ``objectness measure" -- a scalar $[0,1]$ probability of the box $r_j$ having an object. (ii) Using proposals $\mathbf{r} \in \mathbb{R}^{R \times 4}$ we pool the feature map $f_{\mathbf{p}}$ via RoI pool \cite{fast-rcnn} to get intermediate feature of size $\mathbb{R}^{7 \times 7 \times 512}$, followed by a fully-connected layer (FC) to get $f_{\mathbf{r}} \in \mathbb{R}^{R \times 512}$ as, $f_{\mathbf{r}} = \mathcal{P}(f_{\mathbf{p}}, \mathbf{r})$. The patch feature $f_{\mathbf{r}}$ is branched into two streams -- a classification branch $\phi_{\mathrm{cls}}: \mathbb{R}^{R \times 512} \rightarrow \mathbb{R}^{R \times (|\mathcal{C}|+1)}$ outputs probability distribution (per RoI) over $\mathcal{C}$ pre-defined classes and a catch-all background class; a box regressions $\phi_{\mathrm{reg}}: \mathbb{R}^{R \times 512} \rightarrow \mathbb{R}^{R \times 4}$ refines initial box predictions $\mathbf{r} \in \mathbb{R}^{R \times 4}$. 

\keypoint{Baseline SBIR Framework} We recap a baseline SBIR framework. Given a sketch/photo pair ($\mathbf{s}, \mathbf{p}$), we use a sketch/photo feature extractor to get the feature map $f_{\mathbf{s}} = \mathcal{F}_{\mathbf{s}}(\mathbf{s}) \in \mathbb{R}^{512}$ and $f_{\mathbf{p}} = \mathcal{F}_{\mathbf{p}}(\mathbf{p}) \in \mathbb{R}^{512}$. Category-level SBIR requires ($\mathbf{s}, \mathbf{p}$) from the same category, whereas fine-grained SBIR requires instance-level sketch/photo matching. For training, the cosine distance $\delta(\cdot, \cdot)$ to a sketch anchor ($\mathbf{s}$) from a negative photo ($\mathbf{p}^{-}$), denoted as $\beta^{-} = \delta(f_{\mathbf{s}}, f_{\mathbf{p}^{-}})$ should increase while that from the positive photo ($\mathbf{p}^{+}$), $\beta^{+} = \delta(f_{\mathbf{s}}, f_{\mathbf{p}^{+}})$ should decrease. Training is done via triplet loss with hyperparameter $\mu>0$,
\vspace{-0.2cm}
\begin{equation}\label{eq: triplet}
    \mathcal{L}_{\mathrm{trip}} = \max\{0, \mu + \beta^{+} - \beta^{-} \}
\end{equation}
\vspace{-0.6cm}

To extend FG-SBIR across multiple categories (cross-category FG-SBIR), we train with \cref{eq: triplet} using ``hard-triplets" -- $(\mathbf{s}, \mathbf{p}^{+}, \mathbf{p}^{-})$ have same category, and a class discriminator loss across categories using cross-entropy loss,
\vspace{-0.2cm}
\begin{equation}\label{eq: category}
    \mathcal{L}_{\mathrm{cat}} = -c_{\mathbf{q}}^{i} \log \frac{\exp(\mathcal{F}_{\mathbf{c}}(f_{\mathbf{q}}^{i}))}{\sum_{\forall j} \exp(\mathcal{F}_{\mathbf{c}}(f_{\mathbf{q}}^{j}))}
\end{equation}
\vspace{-0.3cm}

\noindent where, query $\mathbf{q} = \{ \mathbf{s}, \mathbf{p} \}$, $c_{\mathbf{q}}^{i}$ represent class label of $i^{th}$ sample, $\mathcal{F}_{\mathbf{c}}: \mathbb{R}^{512} \rightarrow \mathbb{R}^{|\mathcal{C}|}$ predicts softmax probabilities.

\subsection{Weakly Supervised Object Detection}\label{sec: WSOD}
\begin{figure}[!h]
    \centering
    \includegraphics[width=\linewidth]{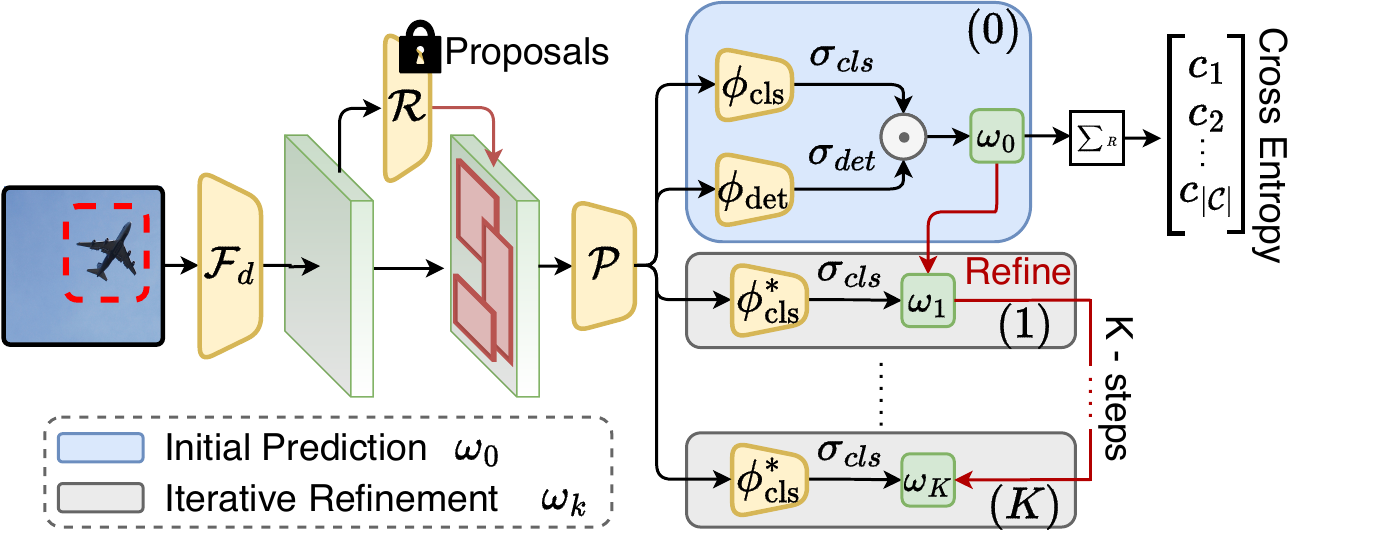}
    \vspace{-0.6cm}
    
    \caption{Weakly supervised setup (no bounding box) trains using image-level class labels with classification heads ($\phi_{\mathrm{cls}}$). Initial prediction $\omega_{0}$ is refined in $K$ steps with $\phi_{\mathrm{cls}}^{*}$ to predict $\omega_{k}$.}
    \label{fig:wsddn}
    \vspace{-0.1cm}
\end{figure}
To avoid collecting expensive bounding box annotation, weakly supervised object detection (WSOD) trains using image-level class labels -- objects of a class are present or not. To avoid using bounding box annotation, we either use a pre-trained region proposal network\footnote{Pre-trained \cite{visualgenome} RPN is highly generalisable to unseen datasets \cite{gu2022open-vocab-OD} due to its generic objective that learns to predict ``objectness" measure.} ($\mathcal{R}$) or heuristic-based selective search \cite{selective-search}, or edge boxes \cite{edgeboxes} that generate box proposals $\mathbf{r} = \{r_1, \dots, r_R\}$. The patch features $f_{\mathbf{r}} = \mathcal{P}(f_{\mathbf{p}}, \mathbf{r})$ is branched into a classification head $x_c = \phi_{\mathrm{cls}}(f_{\mathbf{r}}) \in \mathbb{R}^{R \times (|\mathcal{C}|+1)}$ and a detection head $x_d = \phi_{\mathrm{det}}(f_{\mathbf{r}}) \in \mathbb{R}^{R \times (|\mathcal{C}|+1)}$. The classification head $\phi_{\mathrm{cls}}$ scores individuals proposals into pre-defined $\mathcal{C}$ classes and a catch-all background class via softmax across ($|\mathcal{C}|+1$) class labels
\vspace{-0.3cm}
\begin{equation}
    \sigma_{\mathrm{cls}}(x_c^{(i, j)}) = \frac{ \exp(x_c^{(i, j)}) }{ \sum_{k=1}^{|\mathcal{C}|+1} \exp( x_c^{(i, k)}) }
\end{equation}
The detection head $\phi_{\mathrm{det}}$ measures the contribution of each patch $i$ ($r_i \in \mathbf{r}$) of being classified to class $j$ (in $\mathcal{C}+1$), (i.e., a patch score for each class) via softmax across $R$ regions
\vspace{-0.2cm}
\begin{equation}
    \sigma_{\mathrm{det}}(x_d^{(i, j)}) = \frac{ \exp(x_d^{(i, j)}) }{ \sum_{k=1}^{R} \exp( x_d^{(k, j)}) }
\end{equation}
\vspace{-0.3cm}

We train using image-level labels $\mathrm{\mathbf{Y}} = [y_0, y_1, \dots, y_{|\mathcal{C}|}]^{T} \in \mathbb{R}^{(|C|+1) \times 1}$, where $y_{c} = 1$ or $0$ indicates if instance of class $c \in \mathcal{C}$ is present in the image or not. The combined score (element-wise product) of class score $\sigma_{\mathrm{cls}}$ for each patch and a patch score for each class $\sigma_{\mathrm{det}}$ is computed as, $\omega_{0} = \sigma_{\mathrm{cls}}(x_c) \odot \sigma_{\mathrm{det}}(x_d)$. Since we only have image-level class labels, from the combined score $\omega_{0} \in \mathbb{R}^{R \times (|\mathcal{C}|+1)}$, we take the sum over all patches $\hat{y}_{c} = \sum_{i=1}^{R} \omega_{0}^{i, c}$ to get the probability of instances from the $c^{th}$ class present in the image or not. Training happens via multi-class cross entropy,
\vspace{-0.2cm}
\begin{equation}\label{eq: ws}
    \mathcal{L}_{\mathrm{ws}} = - \sum_{c=1}^{|\mathcal{C}|+1} y_{c} \log \hat{y}_c + (1 - y_c) \log (1 - \hat{y}_c)
\end{equation}
\vspace{-0.4cm}

Unlike SOD, using bounding box annotation to refine proposals, WSOD uses only image-level class labels that fail to naively refine proposals. Hence, we use an iterative refinement classifier $\omega_{k} = \phi_{\mathrm{cls}}^{*}(f_{\mathbf{r}})$, where $\omega_{k} \in \mathbb{R}^{R \times (|\mathcal{C}|+1)}$ to predict a \emph{refined} class score for each RoI, as shown in \cref{fig:wsddn}. The refinement classifier $\phi_{\mathrm{cls}}^{*}$ is supervised via pseudo scores labels ${l}_{k-1}$ from $(k-1)^{th}$ iteration as, (i) we compute the patches with highest scores in each class $r_{*}^{c} = \arg \max_{r} (\omega_{k-1}^{(r, c)})$. (ii) All regions $r_i \in \mathbf{r}$ that has high overlap with a top scoring patch $r_{*}^{c}$ should be the same class label $c$ as, ${l}_{k-1}^{r, c} = 1$ if $\mathrm{IoU}(r_i, r_{*}^{c}) \geq 0.5$. (iii) If a region $r_i \in \mathbf{r}$ has low overlap with any top scoring patch $r_{*}^{c}$, we assign it to background class $l_{k-1}^{r, 0} = 1$. (iv) If a class $c$ is not in image $\mathbf{p}$ we assign $l_{k-1}^{r, c} = 0$. The refinement loss is
\vspace{-0.3cm}
\begin{equation}\label{eq: ref}
    \mathcal{L}_{\mathrm{ref}}^{k} = \frac{1}{R} \sum_{i=1}^{R} \sum_{c=1}^{|\mathcal{C}|} \omega_{k-1}^{(i, j)} \ l_{k-1}^{(i, j)} \ \log \omega_{k}^{(i, j)}
\end{equation}
\vspace{-0.3cm}

Both SOD and WSOD restrict detection to pre-defined $\mathcal{C}$ classes. In the next section, we overcome this fixed-set limitation using prototype learning with SBIR.

\subsection{Localising Object Regions with Query Sketch}
\begin{figure}
    \centering
    \includegraphics[width=\linewidth]{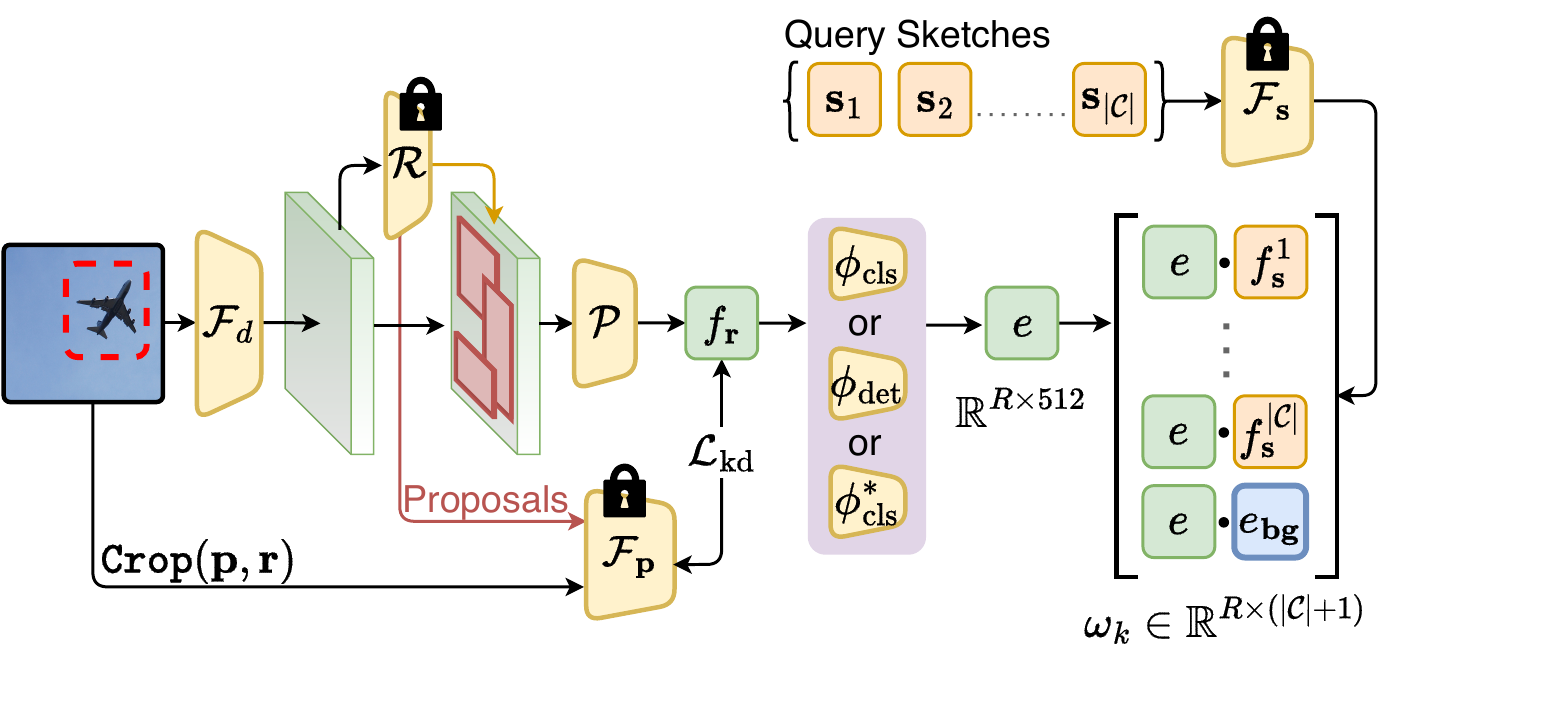}
    \vspace{-1.0cm}
    
    \caption{The object detection modules $\{ \mathcal{F}_{d}, \mathcal{P}, \phi_{cls}, \phi_{det}, \phi_{cls}^{*} \}$ are learned using  pre-trained sketch ($\mathcal{F}_{\mathbf{s}}$) and photo ($\mathcal{F}_{\mathbf{p}}$) encoders.}
    \label{fig:proposed}
    \vspace{-0.3cm}
\end{figure}

We replace the fixed-set classifier in WSOD with scalable open-set prototype learning \cite{prototype-segmentation-2021}. 
Each head $\{\phi_{\mathrm{cls}}, \phi_{\mathrm{det}}, \phi_{\mathrm{cls}}^{*} \}$ in WSOD that predict scores $\mathbb{R}^{R \times 512} \rightarrow \mathbb{R}^{R \times (|\mathcal{C}|+1)}$ is modified to compute their respective embedding vectors $e = \{e_{\mathrm{cls}}, e_{\mathrm{det}}, e_{\mathrm{cls}}^{*} \}$ as $\mathbb{R}^{R \times 512} \rightarrow \mathbb{R}^{R \times 512}$. Next, we compute a support set (prototypes for category-level/instance-level sketch) $\mathcal{S} = [e_{\mathbf{bg}}, f_{\mathbf{s}}^{1}, f_{\mathbf{s}}^{2}, \dots, f_{\mathbf{s}}^{|\mathcal{C}|}]^{T} \in \mathbb{R}^{512 \times (|\mathcal{C}|+1)}$ by encoding query sketches $\{\mathbf{s}_{1}, \dots, \mathbf{s}_{|\mathcal{C}|} \}$ with a pre-trained sketch encoder ($\mathcal{F}_{\mathbf{s}}$) and a learned catch-all background embedding $e_{\mathbf{bg}} \in \mathbb{R}^{512}$, as shown in \cref{fig:proposed}. The scores $\{x_c, x_d, \omega_{k} \}$ (analogous to \cref{sec: WSOD}) are computed using $\mathcal{S}$ and embedding vectors $e$ of detected regions
\vspace{-0.1cm}
\begin{equation}\label{eq: prototype}
    x_c = e_{\mathrm{cls}} \cdot \mathcal{S}; \hspace{1em} x_d = e_{\mathrm{det}} \cdot \mathcal{S}; \hspace{1em} \omega_{k} = e_{\mathrm{cls}}^{*} \cdot \mathcal{S}
\end{equation}
Carefully choosing a sketch encoder $\mathcal{F}_{\mathbf{s}}$ leads to several properties: (i) pre-training $\mathcal{F}_{\mathbf{s}}$ on category-level SBIR computes $\mathcal{S}$ that detect regions $\mathbf{r}$ with the same category as query sketches -- category-level object detection. (ii) pre-training $\mathcal{F}_{\mathbf{s}}$ on cross-category FG-SBIR computes $\mathcal{S}$ where only instance-level aligned regions $\mathbf{r}$ are detected -- fine-grained object detection. (iii) Extending fine-grained object detection with a generalisable (out-of-vocab) sketch encoder $\mathcal{F}_{\mathbf{s}}$ helps to detect object parts (e.g., ``head" of a ``horse") given query sketches -- part-level object detection. We train object detection modules $\{\mathcal{F}_{d}, \mathcal{P}, \phi_{\mathrm{cls}}, \phi_{\mathrm{det}}, \phi_{\mathrm{cls}}^{*}\}$, using \cref{eq: ws} and \cref{eq: ref} in WSOD (\cref{sec: WSOD}).

While the sketch encoder ($\mathcal{F}_{\mathbf{s}}$) trains object detector via prototypes for each category/instance sketch, we further enhance training efficiency with additional supervision from the photo encoder ($\mathcal{F}_{\mathbf{p}}$), as shown in \cref{fig:proposed}. Specifically, we impose a $L1$-based feature matching loss (analogous to feature distillation \cite{heofeaturedistillation2019}) between patch features $f_{\mathbf{r}}$ from proposals $\mathbf{r}$ in object detector and the photo feature computed for cropped photo regions $\texttt{Crop}(\mathbf{p}, \mathbf{r})$ using pre-trained $\mathcal{F}_{\mathbf{p}}$ as, $\mathcal{L}_{\mathrm{kd}} = ||f_{\mathbf{r}} - \mathcal{F}_{\mathbf{p}}( \texttt{Crop}(\mathbf{p}, \mathbf{r}))||_{1}$. The final loss is,
\vspace{-0.2cm}
\begin{equation}\label{eq:knowledge-distillation}
    \mathcal{L}_{\mathrm{tot}} = \underbrace{\mathcal{L}_{\mathrm{ws}} + \sum_{k=1}^{K} \mathcal{L}_{\mathrm{ref}}^{k}}_{\cref{eq: ws} \text{ and } \cref{eq: ref}} + \lambda \underbrace{ ||f_{\mathbf{r}} - \mathcal{F}_{\mathbf{p}}( \texttt{Crop}(\mathbf{p}, \mathbf{r}))||_{1}}_{\mathcal{L}_{\mathrm{kd}}}
\end{equation}
\vspace{-0.3cm}

\noindent where the hyperparameter $\lambda=1$. Although, in theory, we can use our baseline SBIR (in \cref{sec: background}), training object detection requires learning a generalised (out-of-vocab) SBIR for category-level and fine-grained sketch/photo matching under wide variations like illumination, complex background, occlusions, unseen categories etc. 

\subsection{Prompt Learning for Generalised SBIR}\label{sec: prompt-sketch}
To train object detection using SBIR with high generalisation and open-vocab capabilities, we introduce  prompt learning \cite{zhou2022visualprompt} using CLIP \cite{CLIP} for SBIR (both category-level and cross-category fine-grained). CLIP \cite{CLIP} consists an image and text encoder (e.g., ViT \cite{ViT}, or ResNet \cite{he2015resnet}) trained on large $400M$ text/image pairs. This leads to a highly generalisable model that works zero-shot across multiple tasks and datasets. However, adapting CLIP for sketches is tricky since naive fine-tuning leads to model collapse. Hence, we use prompt learning, a set of $P$ learnable vector $\mathbf{v}_{\mathbf{s}} \in \mathbb{R}^{P \times 768}$ for sketch and $\mathbf{v}_{\mathbf{p}} \in \mathbb{R}^{P \times 768}$ for photo, injected into the first layer of ViT to induce CLIP to learn downstream sketch/photo distribution. Importantly, prompting CLIP preserves the desired generalisation ability \cite{zhou2022visualprompt} since the knowledge learned by CLIP is distilled into prompt's weights while keeping the ViT weights frozen. Our new sketch encoder is defined by adapting CLIP's image encoder using sketch prompt ($\mathbf{v}_{\mathbf{s}}$) as, $\mathcal{F}_{\mathbf{s}}(\cdot) = \mathcal{F}_{\mathrm{clip}}(\cdot, \mathbf{v}_{\mathbf{s}})$ and using $\mathbf{v}_{\mathbf{p}}$ for photo encoder as, $\mathcal{F}_{\mathbf{p}}(\cdot) = \mathcal{F}_{\mathrm{clip}}(\cdot, \mathbf{v}_{\mathbf{p}})$. Since ViT weights are frozen, training our CLIP-based SBIR is parameter-efficient -- we train only $\mathbf{v}_{\mathbf{s}} \in \mathbb{R}^{P \times 768}$ and $\mathbf{v}_{\mathbf{p}} \in \mathbb{R}^{P \times 768}$. This allows training with less data, and faster convergence. For category-level SBIR, ($\mathbf{v}_{\mathbf{s}}, \mathbf{v}_{\mathbf{p}}$) learns category inducing prompts using triplet loss (in \cref{eq: triplet}). Learning cross-category FG-SBIR, is slightly more complicated that trains ($\mathbf{v}_{\mathbf{s}}, \mathbf{v}_{\mathbf{p}}$) using hard-triplet in \cref{eq: triplet}, and a modified class discriminative loss \cref{eq: category} using CLIP's text encoder as,
\vspace{-0.1cm}
\begin{equation}
    \mathcal{L}_{\mathrm{cat}} = -c_{\mathbf{q}}^{i} \log \frac{\exp(f_{\mathbf{q}}^{i} \cdot f_{\mathbf{t}}^{i})}{\sum_{\forall j} \exp(f_{\mathbf{q}}^{i} \cdot f_{\mathbf{t}}^{j})}
\end{equation}
\vspace{-0.3cm}

\noindent where, $f_{\mathbf{t}}^{i} \in \mathbb{R}^{512}$ is computed by CLIP's text encoder as, $f_{\mathbf{t}}^{i} = \mathcal{F}_{\mathrm{clip}}^{(\mathbf{t})}(``\texttt{a photo of a }[c_{\mathbf{q}}^{i}]")$ for category $c_{\mathbf{q}}^{i}$. Equipped with our novel prompt-based SBIR, we train open-vocab category-level object detection, fine-grained object detection, and part-level object detection.

\subsection{Bridging Object-Level and Image-Level}\label{sec: object-scene}
\vspace{-0.1cm}

\begin{figure}
    \centering
    \includegraphics[width=\linewidth]{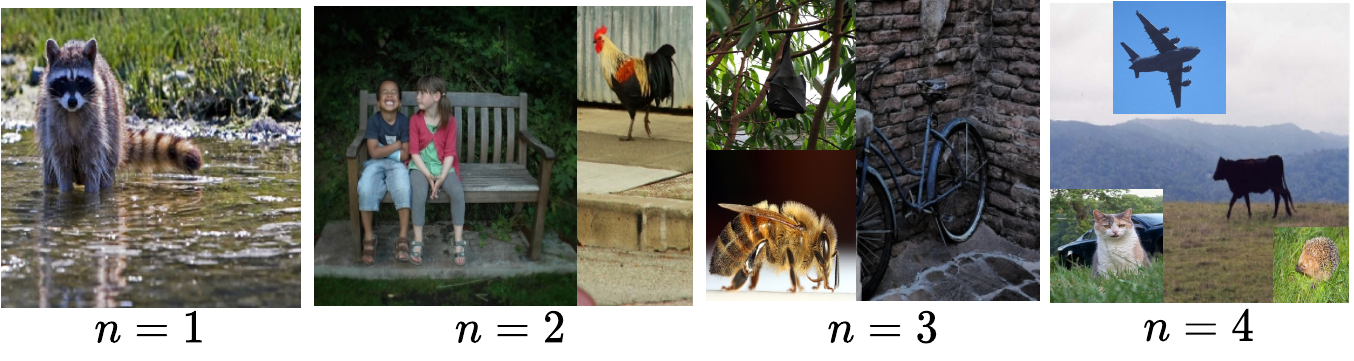}
    \vspace{-0.7cm}
    
    \caption{Bridge object and image-level gap with synthetic photos by tiling $n=\{1, \dots 7\}$ object-level photos in SBIR datasets.}
    \label{fig:data-augmentation}
    \vspace{-0.3cm}
\end{figure}

While SBIR is trained using object-level (single object) sketch/photo pairs, object detection works on image-level (multiple objects) data. To train object detectors using SBIR, we need to bridge this object and image-level gap. Our solution is embarrassingly simple -- synthesise a canvas of size $(H \times W)$ by randomly tiling $n=\{1, \dots, 7\}$ object-level photos in SBIR datasets \cite{sketchrnn2018, sketchy}. Despite its simplicity, our augmentation trick, analogous to CutMix \cite{cutmix}, improves robustness against input corruptions and out-of-distribution generalisation \cite{mixup, cutmix}. The paired sketches for photos in canvas are used to construct the support set $\mathcal{S}$. Note, we train our object detector without the need to ``see" the evaluation data distribution or use any annotation (bounding box or image-level class labels). We call this setup -- extremely weakly supervised object detection (EWSOD) -- no need to ``see" the downstream data distribution.

\section{Experiments}

\keypoint{Dataset} We train our object detector using existing cross-category FG-SBIR dataset -- Sketchy \cite{sketchy} that contains $125$ categories, each with $100$ photos. Every photo in \cite{sketchy} has at least $5$ instance-level paired sketches. To evaluate fine-grained object detection, we use SketchyCOCO \cite{sketchycoco2020} comprising of natural images in MS-COCO \cite{mscoco} with instance-level paired sketches. Following Liu \etal \cite{liu2020scenesketcher}, we select $1,225$ sketch/photo pairs from SketchyCOCO \cite{sketchycoco2020} with at least one foreground sketched object. We filter the overlapping categories of in SketchyCOCO \cite{sketchycoco2020} from Sketchy \cite{sketchy} to measure true zero-shot performance. For category-level object detection, we train on category-level sketch/photo pairs in QuickDraw-Extended \cite{doodle-to-search} having $330k$ sketches and $204k$ photos from $110$ categories. Following \cite{tripathi2020object}, we evaluate on a subset of standard object detection datasets like PASCAL-VOC \cite{pascalVOC} and MS-COCO \cite{mscoco} that have $20$ and $56$ overlapping categories in QuickDraw \cite{sketchrnn2018}.

\keypoint{Implementation Details} Our model is implemented in PyTorch on a 11GB Nvidia RTX 2080-Ti GPU. First, we train a generalised cross-category FG-SBIR with image size ($224 \times 224$) by adapting CLIP with ViT \cite{ViT} backbone (ViT-B/32 weights) using prompt learning \cite{VPT}. The prompts ($P=3$) are trained with triplet loss \cite{yu2016shoe},  margin $\mu=0.3$, Adam optimiser with learning rate $1e-4$ for $60$ epochs, and batch size $64$. Our object detection pipeline is build using Detectron2 \cite{detectron2}. We use FasterRCNN \cite{faster-rcnn}, pretrained on Visual Genome \cite{visualgenome} and remove the RoIPooling \cite{fast-rcnn} and subsequent layers to keep only the pretrained backbone ResNet+FPN ($\mathcal{F}_{d}$) \cite{he2015resnet, lin2017FPN} and Region Proposal Network ($\mathcal{R}$) that generates $1000$ proposals. An alternative is to use handcrafted region proposals like selective search \cite{selective-search}, but we observed slight performance drop. The object detector trains using SGD with batch size $8$ and initial learning rate $5e-3$, multiplied by $0.1$ at $150k$ and $250k$ iterations. We train in a two-step process: (i) keeping $\mathcal{F}_{d}$ and $\mathcal{R}$ fixed, we train the RoI pooling and FC layers ($\mathcal{P}$), classification head ($\phi_{cls}$), detection head ($\phi_{det}$), and refinement head ($\phi_{cls}^{*}$) for $240k$ iterations. (ii) We freeze only $\mathcal{R}$ and finetune all modules for $80k$ iterations. Non-maxima suppression with IoU $\geq 0.3$ is applied to get final predictions.

\keypoint{Evaluation Metric}  (i) For fine-grained object detection, we measure $AP_{.3}$, $AP_{.5}$, and $AP_{.7}$ that computes the average precision (AP) at IoU values $0.3$, $0.5$, and $0.7$. (ii) For category-level object detection, we use measure $AP_{.5}$ and \emph{CorLoc} that computes percentage of images for which the most confident predicted box has IoU $\geq 0.5$ with at least one of the ground-truth boxes for every class. (iii) For cross-category FG-SBIR, we measures $Acc.@q$ -- percentage of sketches having true matched photo in the top-q list, and (iv) mean average precision (mAP), and precision considering top $200$ retrievals P@200 for category-level SBIR.

\subsection{Competitors}  
For object detection, we compare against, (i) supervised object detection (SOD) using both bounding box in addition to sketch/photo annotations: \textbf{Mod-FRCNN} adapts Faster-RCNN \cite{faster-rcnn} for unseen class by concatenating query sketch feature with the RoI pooled feature followed by a binary classifier. \textbf{MatchNet} \cite{matchnet2019} extends \emph{Mod-FRCNN} using co-attention to generate region proposals conditioned on query sketch along with squeeze-and-co-excitation to adaptively re-weight importance distribution of candidate proposals. \textbf{CoAttOD} \cite{tripathi2020object} improves upon \emph{MatchNet} by mitigating the sketch/photo domain misalignment using cross-modal attention. (ii) Weakly supervised object detection (WSOD) trains only on image-level sketch annotations without any additional bounding boxes: \textbf{WSDDN} \cite{wsddn} repurposed object detection as a region classification via multiple instance learning (MIL) paradigm. To inject query sketch to \emph{WSDDN}, we use cross-attention with RoI pooled feature followed by a binary classifier for detection. \textbf{OICR} \cite{OICR} improves \emph{WSDDN} with an iterative MIL to refine initial predictions scores to improve discriminatory power for detection. \textbf{PCL} \cite{pcl2018} generates multiple positive instance in an image via clustering and assigning proposals to the label of corresponding object class for each cluster. \textbf{ICMWSD} \cite{ren2020WSOD} addresses the problem of prior WSOD that focus on the most discriminative part of an object using context information. In particular, \emph{ICMWSD} obtains a ``dropped feature" by dropping the most discriminative parts, followed by maximising the loss of the ``dropped feature" that force the network to look in the surrounding context regions. (iii) We adapt \textbf{$<$Method$>$} in \emph{WSOD} to \textbf{E-$<$Method$>$} that exclusively training on SBIR datasets \cite{sketchrnn2018, sketchy} by synthesising canvas with randomly tiling $n=\{1, \dots, 7\}$ object-level photos and using their paired sketches to construct the support $\mathcal{S}$. We call this setup -- extreme weakly supervised object detection (EWSOD).

For zero-shot category-level SBIR, we compare against: \textbf{GRL} \cite{doodle-to-search} combines similar semantic information (word2vec \cite{word2vec2013}) of class labels with visual sketch information and trains using a gradient reversal layer \cite{grl2015} to reduce sketch/photo domain gap. \textbf{VKD} \cite{wang2022ViTKD} is similar to ours using prototype-learning but employ selective knowledge distillation and ViT \cite{ViT} backbone. For zero-shot cross-category FG-SBIR: \textbf{CDG} is a SOTA domain generalisation method \cite{shankar2018} adapted to cross-category FG-SBIR \cite{pang2019generalising} using categories as domain and intra-category sketch/photo pairs as label. \textbf{CCD} \cite{pang2019generalising} models a universal manifold of prototypical visual sketch traits that dynamically embeds sketch/photo, to generalise to unseen categories. 

\subsection{Generalisibility of Cross-Category FG-SBIR} 
Due to the significant impact of SBIR on training object detectors, it is imperative to learn a powerful cross-category FG-SBIR that is highly generalisable. In other words, the accuracy of SBIR puts a bottleneck on object detection performance. \cref{tab:generalised-SBIR} compares category-level SBIR (CL-SBIR) and cross-category FG-SBIR (CC-FGSBIR) on QuickDraw-Extended \cite{doodle-to-search} and Sketchy \cite{sketchy} respectively, using $100\%$, $70\%$, and $50\%$ of the training set.

\keypoint{Performance Analysis} From \cref{tab:generalised-SBIR} we make the following observations: (i) with decreasing train-set categories, the performance gap (ratio of proposed / SOTA) between the proposed method versus GRL (for CL-SBIR) and CDG (for CC-FGSBIR) increases from $2.1/1.4$ at $100\%$ data to $3.0/4.2$ at $50\%$ data. This shows the high generalisation potential when using prompt-based CLIP models for sketch/photo matching. (ii) Performance gap of proposed versus SOTAs for $100\% \rightarrow 50\%$ is more significant in CC-FGSBIR as compared to CL-SBIR. Hence, it is more difficult to discriminate unseen intra-category sketch/photo pairs than recognise a novel categories. (iii) Performance of all competitors in CL-SBIR and CC-FGSBIR are staggeringly inferior to proposed CLIP-based approach. Such a strong SBIR is necessary to unlock training object detection in EWSOD setup (cross-dataset and weakly supervised).

{
\setlength{\tabcolsep}{0.6em}
\begin{table}[]
    \centering
    \footnotesize
    \caption{Quantitative performance of zero-shot category-level SBIR (CL-SBIR) and cross-category FG-SBIR (CC-FGSBIR).}
    \vspace{-1em}
    \begin{tabular}{ccccccc}
        \toprule
        \multirow{2}{*}{\makecell{Train}} &  & \multicolumn{2}{c}{CL-SBIR \cite{doodle-to-search}} & & \multicolumn{2}{c}{CC-FGSBIR \cite{sketchy}} \\
         &  & mAP & P@200 & & Acc.@1 & Acc.@5 \\\hline
        \multirow{3}{*}{$100\%$} & GRL & 9.01 & 6.75 & CDG & 20.1 & 46.4 \\
         & VKD & 15.0 & 29.8 & CCD & 22.6 & 49.0 \\
        \rowcolor{Gray}
         & \multicolumn{1}{c}{\textbf{Ours}} & \textbf{18.2} & \textbf{36.1} & \textbf{Ours} & \textbf{27.6} & \textbf{59.5} \\\hline
        \multirow{3}{*}{$70\%$} & GRL & 6.3 & 5.7 & CDG & 14.6 & 39.5 \\
         & VKD & 9.4 & 17.3 & CCD & 16.3 & 41.4 \\
        \rowcolor{Gray}
         & \multicolumn{1}{c}{\textbf{Ours}} & \textbf{13.1} & \textbf{23.2} & \textbf{Ours} & \textbf{21.0} & \textbf{47.7} \\\hline
        \multirow{3}{*}{$50\%$} & GRL & 3.2 & 2.7 & CDG & 7.9 & 25.4 \\
         & VKD & 4.8 & 6.3 & CCD & 9.2 & 32.2 \\
         \rowcolor{Gray}
         & \multicolumn{1}{c}{\textbf{Ours}} & \textbf{9.6} & \textbf{11.4} & \textbf{Ours} & \textbf{14.7} & \textbf{40.1} \\\bottomrule
    \end{tabular}
    \label{tab:generalised-SBIR}
\end{table}
}

\subsection{Category-Level Object Detection}

\begin{figure}
    \centering
    \includegraphics[width=\linewidth]{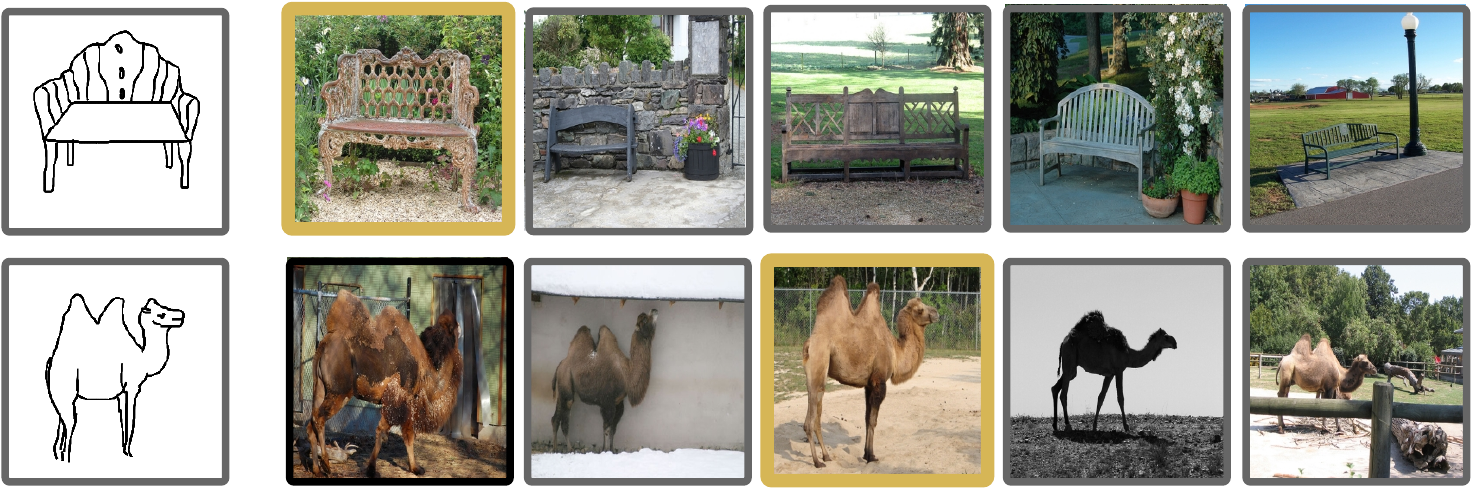}
    \vspace{-2em}
    
    \caption{Qualitative retrieval results for cross-category FG-SBIR.}
    \label{fig: retrieval-cross-category}
    \vspace{-0.4cm}
\end{figure}

We benchmark on a subset of standard object detection PASCAL-VOC \cite{pascalVOC} and MS-COCO \cite{mscoco} datasets that have overlapping categories with QuickDraw \cite{sketchrnn2018} sketches. Unlike traditional object detection that detects all instances for known classes in an image, category-level object detection specifies the category of interest by drawing a query sketch.

\keypoint{Performance Analysis} From \cref{tab: detection-category} we observe: (i) best SOD method outperform the best WSOD by an average $AP_{.5}$ margin of $1.7\%/0.1\%$ in VOC/MS-COCO. This shows that although WSOD performs less than SOD (using additional bounding box annotation), the performance gap is not as significant as generally observed in prior works on seen setup using text as query \cite{gu2022open-vocab-OD, OICR, ren2020WSOD}. In other words, using sketch gives nearly similar performance for zero-shot setup for SOD and WSOD. (ii) EWSOD methods further drops $AP_{.5}$ of best WSOD method by $5.1\%/1.6\%$. This highlights the lack of generalisation of object detectors to the shift in data distribution when trained on SBIR photos and tested on VOC/MS-COCO. (iii) Despite being trained on the challenging EWSOD setup, our proposed method outperforms best SOD by $14.7/10.9$, WSOD by $16.4/11.0$, and EWSOD by $21.5/13.2$ in zero-shot setup. This shows the extreme generalisation potential of training object detetction using a strong CLIP-based SBIR.

{
\setlength{\tabcolsep}{0.8em}
\begin{table}[]
 \centering
 \footnotesize
 \caption{Quantitative performance of category-level object detection on VOC 2007 and MS-COCO using $AP_{.5}$ and \emph{CorLoc}.}
 \vspace{-1em}
 \begin{tabular}{clcccc}
     \toprule
     \multicolumn{2}{c}{\multirow{2}{*}{Method}} & \multicolumn{2}{c}{VOC 2007 \cite{pascalVOC}} & \multicolumn{2}{c}{MS-COCO \cite{mscoco}} \\
      & & $AP_{.5}$ & CorLoc & $AP_{.5}$ & CorLoc \\\hline
     \multirow{3}{*}{\rotatebox{90}{\textbf{SOD}}} & Mod-FRCNN & 30.1 & 51.2 & 7.4 & 65.8 \\
      & MatchNet & 31.4 & 51.7 & 12.4 & 68.1 \\
      & CoAttOD & 34.6 & 53.9 & 15.0 & \textbf{71.3} \\\hline
     \multirow{4}{*}{\rotatebox{90}{\textbf{WSOD}}} & WSDDN & 20.9 & 40.1 & 11.9 & 67.3 \\
      & OICR & 24.7 & 42.3 & 12.2 & 67.7 \\
      & PCL & 26.1 & 45.5 & 13.8 & 68.6 \\
      & ICMWSD & 32.9 & 52.6 & 14.9 & 69.5 \\\hline
     \multirow{4}{*}{\rotatebox{90}{\textbf{EWSOD}}} & E-WSDDN & 17.7 & 37.9 & 10.1 & 66.7 \\
      & E-OICR & 21.2 & 40.5 & 10.4 & 67.0 \\
      & E-PCL & 22.3 & 41.1 & 11.8 & 67.3 \\
      & E-ICMWSD & 27.8 & 46.3 & 12.7 & 67.9 \\
     \rowcolor{Gray}
      & {\textbf{Proposed}} & \textbf{49.3} & \textbf{69.4} & \textbf{25.9} & {70.3} \\\bottomrule
 \end{tabular}
 \label{tab: detection-category}
\end{table}
}

\begin{figure}
    \centering
    \includegraphics[width=\linewidth]{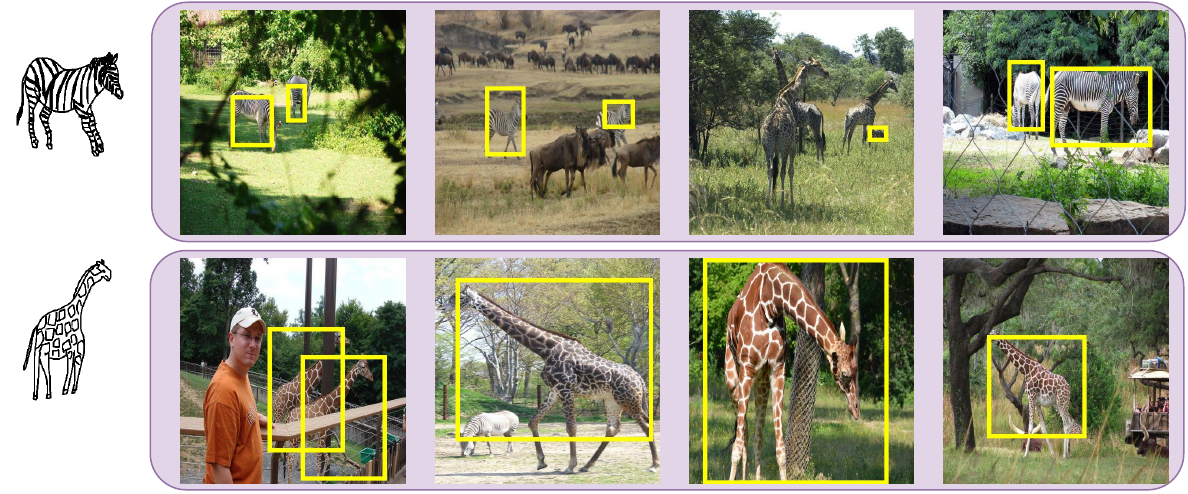}
    \vspace{-2em}
    
    \caption{Category-Level Object Detection using query sketches with images from MS-COCO \cite{mscoco} and PASCAL-VOC \cite{pascalVOC}.}
    \label{fig: detection-category}
    \vspace{-0.4cm}
\end{figure}
 
\subsection{Fine-Grained Object Detection}
Unlike category-level object detection that detects all instances of sketched category, the goal of fine-grained object detection is to detect only a specific instance for the input query sketch with instance-level alignment.

\keypoint{Performance Analysis} From \cref{tab: detection-finegrained} we observe: (i) Methods in SOD have nearly similar performance as WSOD and drops for EWSOD, similar to that in category-level detection in \cref{tab: detection-category}. (ii) Compared to SOD, the performance of WSOD drops more for $AP_{.5} \rightarrow AP_{.7}$. This is since WSOD methods use less accurate selective search \cite{selective-search} and edge boxes \cite{edgeboxes} for region proposals compared to the more accurate RPN \cite{faster-rcnn} in SOD. (iii) Our proposed method outperforms SOD, WSOD, and EWSOD in zero-shot setup, thereby proving its fine-grained generalisation.
 
{
\setlength{\tabcolsep}{1.2em}
\begin{table}[]
    \centering
    \footnotesize
    \caption{SketchyCOCO detection fine-grained.}
    \vspace{-1em}
    \begin{tabular}{clccc}
        \toprule
        & Method & $AP_{.3}$ & $AP_{.5}$ & $AP_{.7}$ \\\hline
         \multirow{3}{*}{\rotatebox{90}{\textbf{SOD}}} & Mod-FRCNN & 2.5 & 3.5 & 3.1 \\
          & MatchNet & 9.3 & 11.0 & 10.5 \\
          & CoAttOD & 10.4 & 12.1 & 11.7 \\\hline
         \multirow{4}{*}{\rotatebox{90}{\textbf{WSOD}}} & WSDDN & 8.1 & 10.2 & 9.4 \\
          & OICR & 8.9 & 10.9 & 10.0 \\
          & PCL & 9.2 & 11.5 & 10.6 \\
          & ICMWSD & 10.3 & 11.9 & 10.8 \\\hline
         \multirow{4}{*}{\rotatebox{90}{\textbf{EWSOD}}} & E-WSDDN & 6.4 & 8.5 & 7.6 \\
          & E-OICR & 7.1 & 9.1 & 8.3 \\
          & E-PCL & 7.3 & 9.4 & 8.7 \\
          & E-ICMWSD & 8.5 & 10.2 & 9.4 \\
         \rowcolor{Gray}
          & {\textbf{Proposed}} & \textbf{15.0} & \textbf{17.1} & \textbf{16.3} \\\bottomrule
    \end{tabular}
    \label{tab: detection-finegrained}
    \vspace{-0.1em}
\end{table}
}
 
 \begin{figure}
    \centering
    \includegraphics[width=\linewidth]{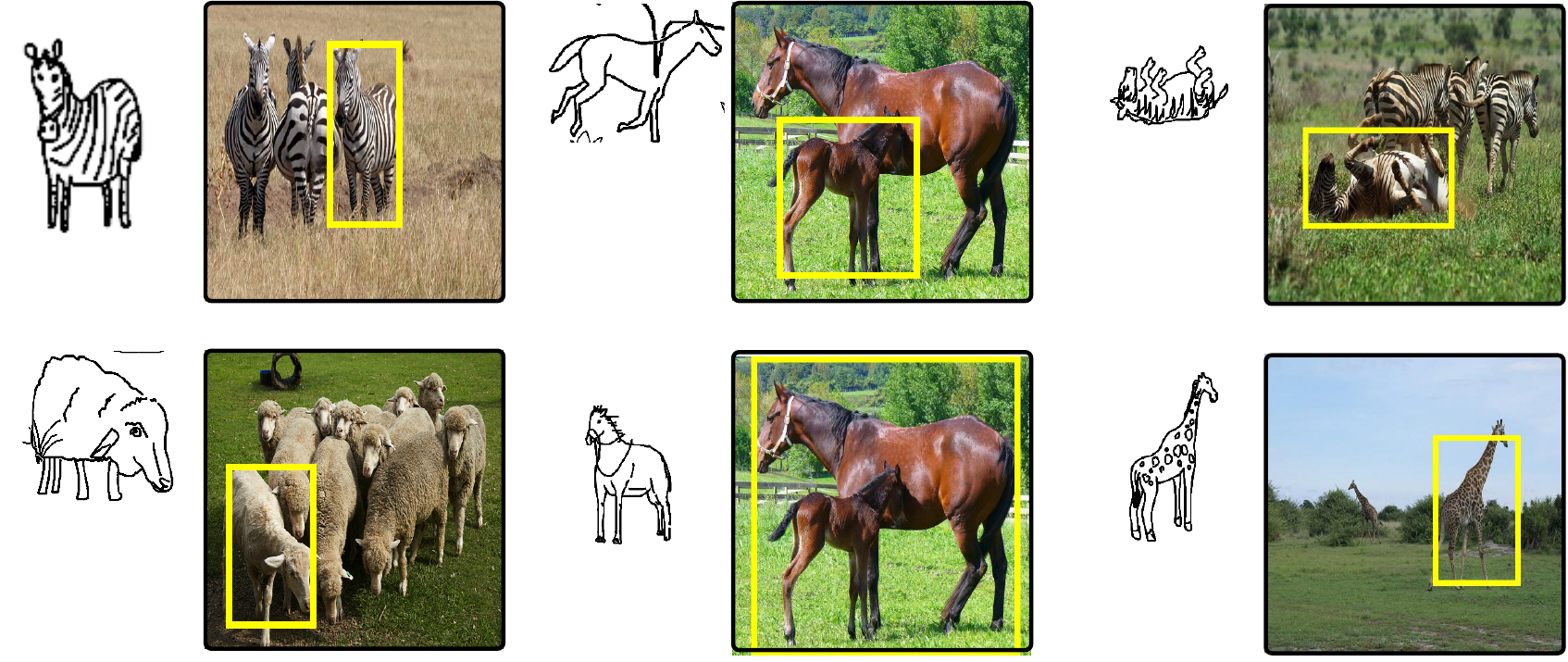}
    \vspace{-2.2em}
    
    \caption{Cross-Category Fine-Grained Object Detection using query sketches with images from SketchyCOCO \cite{sketchycoco2020}.}
    \label{fig: detection-finegrained}
    \vspace{-0.5cm}
\end{figure}
 
\subsection{Part-Level Object Detection}
Encouraged with the generalised fine-grained discriminative power of the proposed method in \cref{tab: detection-finegrained}, we go a step further and ask: can we only detect a \emph{part} (e.g., only `head') of an instance? Due to lack of annotation, a quantitative evaluation of part-level object detection is infeasible. Nonetheless, we conduct a qualitative study by manually editing sketches to create partial sketches of a single part (e.g., only ``head" of ``horse"). \cref{fig: detection-part} presents some results (for more see supplementary). We observe that (i) our proposed method can uniquely detect the sketched `head' region of different objects. (ii) Detection performance is lower for ambiguous part sketches like `leg' (e.g., front-leg, back-leg etc.) (iii) Since detection depends on region proposals from RPN, our model fails to detect tiny sketched parts. Tiny object detection \cite{lee2022tinyOD} is a known challenge for traditional object detection \cite{faster-rcnn}.

\begin{figure}
    \centering
    \includegraphics[width=\linewidth, height=0.5\linewidth]{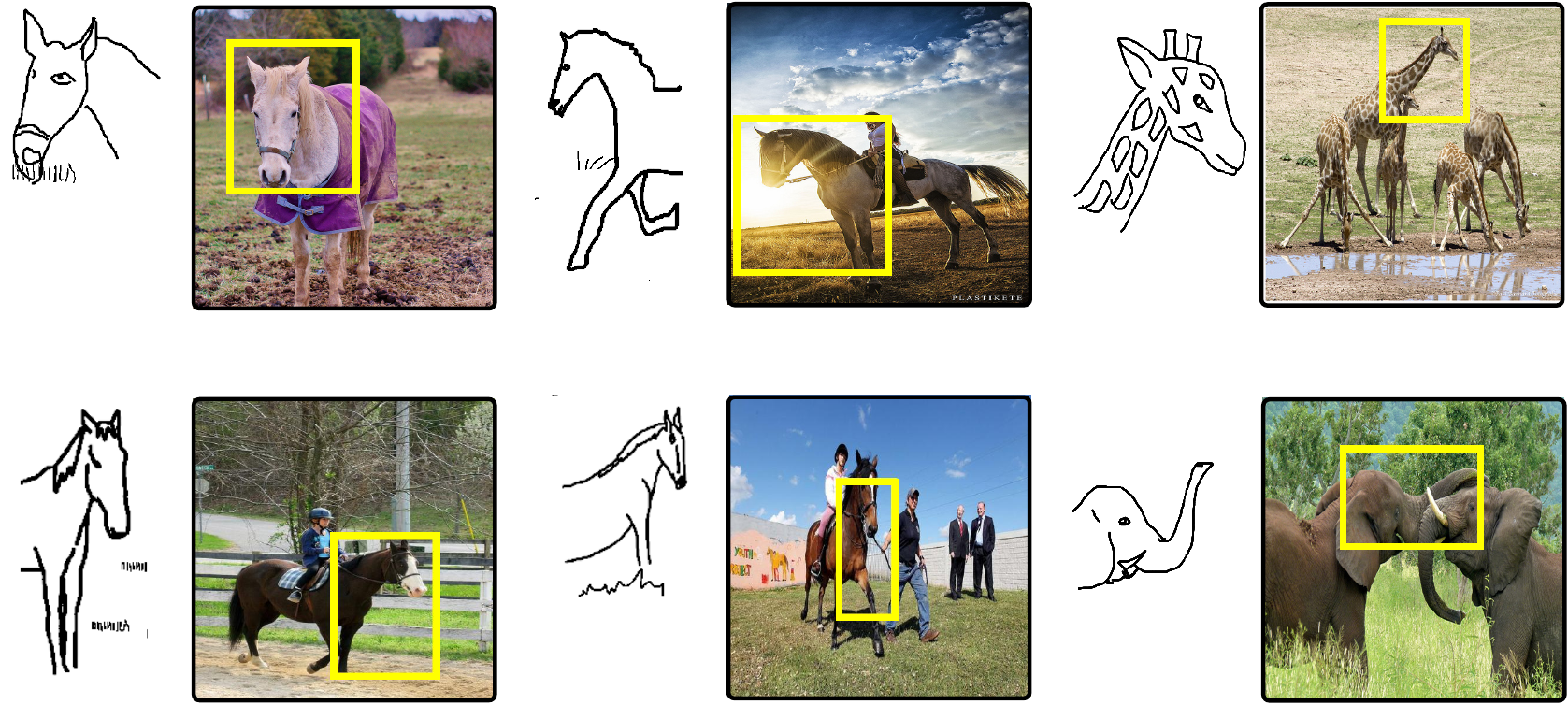}
    \vspace{-2em}
    
    \caption{Unlike traditional object detection that detects an entire object (e.g., a ``horse"), sketches can express fine-grained RoI to detect a specified part of an object (e.g., the ``head" of a ``horse").}
    \label{fig: detection-part}
    \vspace{-0.5cm}
\end{figure}

\subsection{Ablation}
\keypoint{Selective Search v/s Edge Boxes v/s RPN} Unlike the proposed method using pre-trained \cite{visualgenome} RPN to generate $1000$ box proposals, WSOD methods mostly use selective search \cite{selective-search} (SS) or edge boxes \cite{edgeboxes} (EB) that do not need pre-training using box annotation from visual genome \cite{visualgenome}. Hence, for a fair comparison, we replace RPN with SS/EB drops $AP_{.5}$ performance by $1.3$/$2.6$ on SketchyCOCO \cite{sketchycoco2020}.

\keypoint{Influence of Classifier Refinement} We observe $AP_{.5}$ improve by $2.4$ and $1.1$ for $K=1 \rightarrow 2$ and $K=2 \rightarrow 3$ respectively but a small drop of $0.2$ for $K=3 \rightarrow 4$. 

\keypoint{Influence of Supervision from Photo Encoder in SBIR} Although we can train an object detector using only pre-trained sketch encoder (trained on SBIR) via prototype learning, removing supervision from the photo encoder in SBIR drops $4.5$ in $A_{.5}$ on SketchyCOCO \cite{sketchycoco2020}.
 
\subsection{Limitation and Future Works} 
Introducing fine-grained object detection using sketch opens several possibilities that we do not consider. Given multiple query sketches, currently we tread them as independent query embeddings. However, a user might be interested in detecting complex scenes (a ``dog" on the right of a ``person") with \emph{multiple} objects that have meaningful \emph{spatial} alignment. Future works can extend fine-grained object detection to semantic segmentation using complex sketches from the recently introduced FS-COCO \cite{fscoco} dataset.
\vspace{-0.1em}

\section{Conclusion}

We cultivate the expressiveness that human sketch bring for object detection. The proposed sketch-enabled object detection framework detects what \emph{you} intend to express in \emph{your} sketch -- an object detector that is both instance-aware and part-aware. Accordingly, we design a novel prompt learning setup to marry CLIP and SBIR, to train a sketch-aware detector, that works without needing bounding box annotation, or class labels. To make our detector general-purpose, we further dictate it to work in a zero-shot manner. While SBIR is trained using object-level (single object) sketch/photo pairs, object works on image-level (multiple categories). We bridge this object and image-level gap using a data augmentation trick that improves robustness towards corruption and generalisation to out-of-vocab. The resulting framework outperforms both supervised, and weakly supervised object detectors on zero-shot setup.

\appendix

\section{Human Study on Part-level Object Detection}
Due to the lack of annotation, a quantitative evaluation of part-level object detection is infeasible. Nonetheless, we measure the real-world usability of our sketch-enabled object detection framework using Mean Opinion Score (MOS) by asking $10$ people to draw $20$ part-level sketches and rate from $1$ to $5$ (bad $\rightarrow$ excellent) based on their opinion of how closely the queried object part was detected. Accordingly, we obtain a MOS (mean $\pm$ variance of $200$ responses) of $3.67 \pm 0.6$. 

\section{Preliminary Study on Occluded Objects}
In addition to category-level, fine-grained, and part-level object detection, we further qualitatively test the generalisability of the system to detect occluded objects as:
\begin{figure}[!h]
    \centering
    \includegraphics[width=\linewidth]{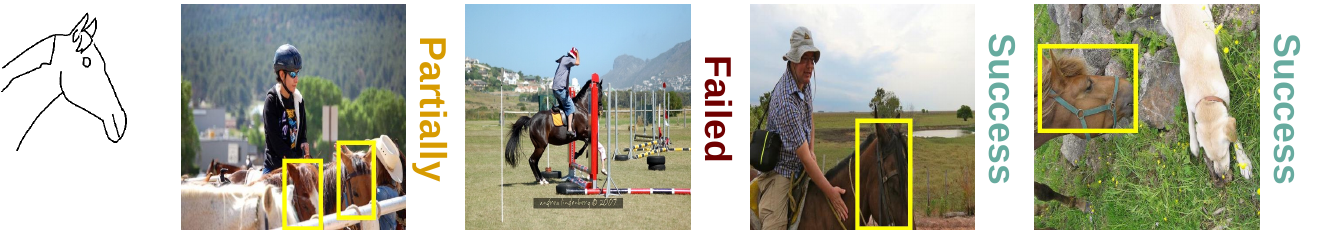}
    \label{fig:my_label}
    \vspace{-0.5cm}
\end{figure}

While we show some successful, failed, and partially detected cases, future works can further investigate the role of sketch and foundation models like CLIP \cite{CLIP} for occluded object detection.

\section{Relation to Open World setup}
In open world setup, a model trained on $C$ known classes can recognise the unknown class and update the base model via incremental learning \cite{bendale2015openWorld, joseph2021OW}. Our method already works in open world setup as it detects in zero-shot, open-vocab setup, i.e., it works regardless of whether the query sketch is in the train set or not.

\section{Detection across Different Poses}
Our object detection has multiple setups: (i) for category-level OD, the sketch of object O1 (``zebra") in image I1 will detect the same object O1 in a different image I2 \emph{even with a different pose} (``sitting" or ``standing"). (ii) For fine-grained OD, the sketch of object O1 in image I1 will only detect the same object O1 in a different image I2 if it has the \emph{same pose}, e.g., detect only ``zebras \textit{sitting down}" amongst a herd of ``zebras".  Figure below shows qualitative results for clarity.
\includegraphics[width=\linewidth, height=0.2\linewidth]{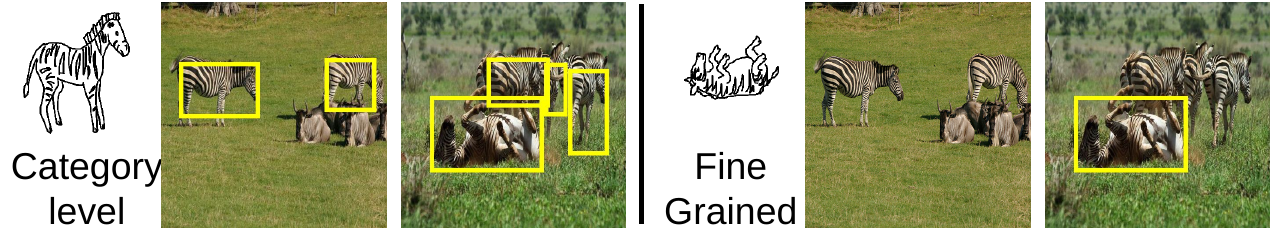}

\section{Additional Ablation Study}
(i) \textit{Varying prompt length} $P=\{1, 3, 5\}$ in $\{\mathbf{v}_{\mathbf{s}}, \mathbf{v}_{\mathbf{p}} \} \in \mathbb{R}^{P \times 768}$ changes $AP_{.5}$ to $16.5$, $17.1$, and $15.9$ on SketchyCOCO~\cite{sketchycoco2020} respectively. (ii) \textit{Replacing CLIP} with VGG-based sketch encoder $\mathcal{F}_{\mathbf{s}}$ sharply drops $AP_{.5}$ to $9.1$ (iii) \textit{Increasing tiling} from $n\in[1,7]$ to $n\in[1,17]$ reduces $AP_{.5}$ to $11.3$ due to high occlusion ($n\rightarrow17$).

\section{Robustness to Tiling}
To test robustness, we generate occluded photos by randomly masking $(10\%, 30\%, 50\%)$ of GT object boundaries with zero pixel values and measure the respective drop in accuracy ($AP_{.5})$ on \cite{sketchycoco2020}. Performance drop being less \textit{with tiling} for E-WSDDN (by $\{1.7, 3.4, 5.7\}$) or our method (by $\{1.6, 3.3, 5.4\}$), than \textit{without tiling} in WSDDN (by $\{3.1, 5.2, 7.5\}$) verifies robustness due to tiling on object detection.

\section{Clarification on CutMix \cite{cutmix}  vs. our Tiling}
(i) Our novelty lies in adapting well-known modules (CLIP, SBIR) to train an object detector from only object-level sketch-photo pairs (each photo has only one object) without any bounding-box annotations. (ii) Despite sharing a common technical implementation, CutMix \cite{cutmix} is a data \textit{augmentation} tool that typically replaces a patch in one \textit{existing} scene-photo with that from another. Contrarily, tiling is a data \textit{synthesis} tool that combines multiple object-level photos in the SBIR dataset to \textit{newly create} a scene photo for subsequent training.

\begin{figure*}
    \centering
    \includegraphics[width=\linewidth]{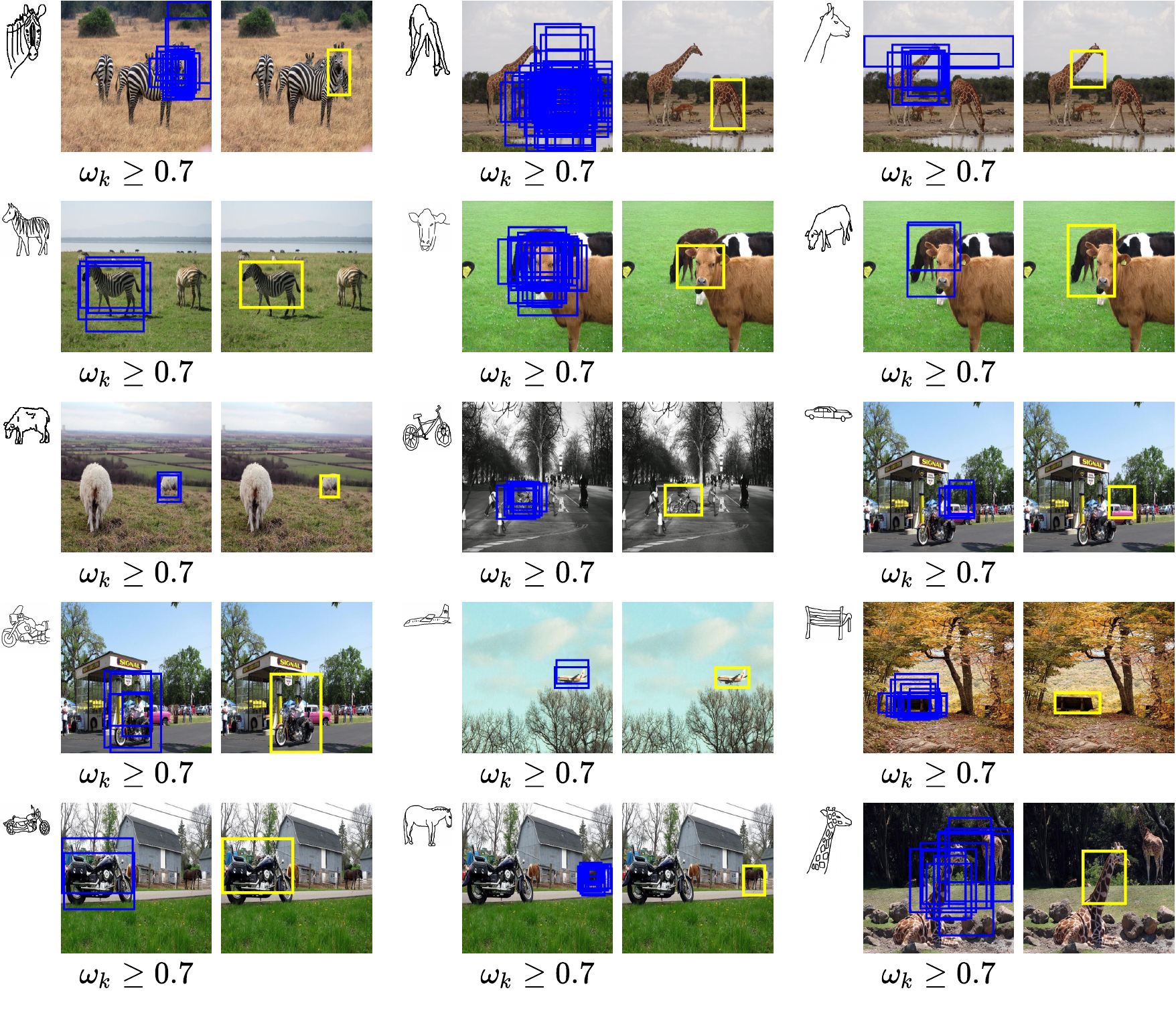}
    \caption{Additional qualitative results for fine-grained and part-level object detection on SketchyCOCO. Note both the \textcolor{blue}{Blue} and \textcolor{yellow}{Yellow} boxes are network predictions and not ground truth. The \textcolor{blue}{Blue} boxes are predictions from the network prior to using Non-maximum suppression (NMS) with the confidence score of the predicted box $\omega_{k} \geq 0.7$. The \textcolor{yellow}{Yellow} boxes are the resulting predictions after applying NMS with $\mathrm{IoU} \geq 0.3$}
    \label{fig:my_label1}
\end{figure*}

{\small
\bibliographystyle{ieee_fullname}
\bibliography{egbib}
}

\end{document}


\RestyleAlgo{ruled}
\title{What Can Human Sketches Do for Object Detection? \\ Supplemental material\vspace{-0.5cm}}

\lstset{
    basicstyle=\ttfamily,
    keywordstyle=\color{blue},
    stringstyle=\color{DarkMagenta},
    commentstyle=\color{DarkGreen},
    morecomment=[l]{\%}
}

\author{Pinaki Nath Chowdhury \hspace{.2cm}  Ayan Kumar Bhunia \hspace{.2cm} Aneeshan Sain \hspace{.2cm}  Subhadeep Koley \\
Tao Xiang\hspace{.3cm}  Yi-Zhe Song \\
SketchX, CVSSP, University of Surrey, United Kingdom.  \\
{\tt\small \{p.chowdhury, a.bhunia, a.sain, s.koley, t.xiang, y.song\}@surrey.ac.uk} 
}
\maketitle


\renewcommand{\thefigure}{\Alph{figure}}
\renewcommand{\thesubsection}{\Alph{subsection}}

\definecolor{commentcolor}{RGB}{110,154,155}   
\newcommand{\PyComment}[1]{\ttfamily\textcolor{commentcolor}{\# #1}}  
\newcommand{\PyCode}[1]{\ttfamily\textcolor{black}{#1}} 

\subsection{Human Study on Part-level Object Detection}
Due to the lack of annotation, a quantitative evaluation of part-level object detection is infeasible. Nonetheless, we measure the real-world usability of our sketch-enabled object detection framework using Mean Opinion Score (MOS) by asking $10$ people to draw $20$ part-level sketches and rate from $1$ to $5$ (bad $\rightarrow$ excellent) based on their opinion of how closely the queried object part was detected. Accordingly, we obtain a MOS (mean $\pm$ variance of $200$ responses) of $3.67 \pm 0.6$. 

\subsection{Preliminary Study on Occluded Objects}
In addition to category-level, fine-grained, and part-level object detection, we further qualitatively test the generalisability of the system to detect occluded objects as:
\begin{figure}[!h]
    \centering
    \includegraphics[width=\linewidth]{latex/figures/occlusion.pdf}
    \label{fig:my_label}
    \vspace{-0.5cm}
\end{figure}

While we show some successful, failed, and partially detected cases, future works can further investigate the role of sketch and foundation models like CLIP \cite{CLIP} for occluded object detection.

\subsection{Relation to Open World setup}
In open world setup, a model trained on $C$ known classes can recognise the unknown class and update the base model via incremental learning \cite{bendale2015openWorld, joseph2021OW}. Our method already works in open world setup as it detects in zero-shot, open-vocab setup, i.e., it works regardless of whether the query sketch is in the train set or not.

\subsection{Detection across Different Poses}
Our object detection has multiple setups: (i) for category-level OD, the sketch of object O1 (``zebra") in image I1 will detect the same object O1 in a different image I2 \emph{even with a different pose} (``sitting" or ``standing"). (ii) For fine-grained OD, the sketch of object O1 in image I1 will only detect the same object O1 in a different image I2 if it has the \emph{same pose}, e.g., detect only ``zebras \textit{sitting down}" amongst a herd of ``zebras".  Figure below shows qualitative results for clarity.
\includegraphics[width=\linewidth, height=0.2\linewidth]{latex/figures/category-vs-finegrained-OD.pdf}

\subsection{Additional Ablation Study}
(i) \textit{Varying prompt length} $P=\{1, 3, 5\}$ in $\{\mathbf{v}_{\mathbf{s}}, \mathbf{v}_{\mathbf{p}} \} \in \mathbb{R}^{P \times 768}$ changes $AP_{.5}$ to $16.5$, $17.1$, and $15.9$ on SketchyCOCO~\cite{sketchycoco2020} respectively. (ii) \textit{Replacing CLIP} with VGG-based sketch encoder $\mathcal{F}_{\mathbf{s}}$ sharply drops $AP_{.5}$ to $9.1$ (iii) \textit{Increasing tiling} from $n\in[1,7]$ to $n\in[1,17]$ reduces $AP_{.5}$ to $11.3$ due to high occlusion ($n\rightarrow17$).

\subsection{Robustness to Tiling}
To test robustness, we generate occluded photos by randomly masking $(10\%, 30\%, 50\%)$ of GT object boundaries with zero pixel values and measure the respective drop in accuracy ($AP_{.5})$ on \cite{sketchycoco2020}. Performance drop being less \textit{with tiling} for E-WSDDN (by $\{1.7, 3.4, 5.7\}$) or our method (by $\{1.6, 3.3, 5.4\}$), than \textit{without tiling} in WSDDN (by $\{3.1, 5.2, 7.5\}$) verifies robustness due to tiling on object detection.

\subsection{Clarification on CutMix \cite{cutmix}  vs. our Tiling}
(i) Our novelty lies in adapting well-known modules (CLIP, SBIR) to train an object detector from only object-level sketch-photo pairs (each photo has only one object) without any bounding-box annotations. (ii) Despite sharing a common technical implementation, CutMix \cite{cutmix} is a data \textit{augmentation} tool that typically replaces a patch in one \textit{existing} scene-photo with that from another. Contrarily, tiling is a data \textit{synthesis} tool that combines multiple object-level photos in the SBIR dataset to \textit{newly create} a scene photo for subsequent training.

\begin{figure*}[]
    \centering
    \includegraphics[width=\linewidth]{latex/figures/detection-supp.pdf}
    \caption{Additional qualitative results for fine-grained and part-level object detection on SketchyCOCO. Note both the \textcolor{blue}{Blue} and \textcolor{yellow}{Yellow} boxes are network predictions and not ground truth. The \textcolor{blue}{Blue} boxes are predictions from the network prior to using Non-maximum suppression (NMS) with the confidence score of the predicted box $\omega_{k} \geq 0.7$. The \textcolor{yellow}{Yellow} boxes are the resulting predictions after applying NMS with $\mathrm{IoU} \geq 0.3$}
    \label{fig:my_label}
\end{figure*}


{\small
\bibliographystyle{ieee_fullname}
\bibliography{egbib}
}